\documentclass[runningheads]{llncs}
\usepackage[T1]{fontenc}
\usepackage{graphicx}
\usepackage{booktabs}
\usepackage[misc]{ifsym}
\newcommand{\corr}{(\Letter)}
\usepackage{mwe}
\usepackage{amssymb}

\begin{document}

\title{Distilling Vision Transformers for Distortion-Robust Representation Learning}

\titlerunning{Distilling Vision Transformers for Distortion Robustness}

\author{Konstantinos Alexis\inst{1,3} \corr \and
Giorgos Giannopoulos\inst{2} \and
Dimitrios Gunopulos\inst{1}}

\authorrunning{K. Alexis et al.}

\institute{Department of Informatics and Telecommunications, National and Kapodistrian University of Athens, Greece \email{\{kalexis,dg\}@di.uoa.gr}
\and
National Observatory of Athens, Greece \email{giannopoulos@noa.gr}
\and
Information Management Systems Institute, Athena Research Center, Greece
}

\maketitle              

\begin{abstract}
    Self-supervised learning has achieved remarkable success in learning visual representations from clean data, yet remains challenging when clean observations are sparse or not available at all.
    In this paper, we demonstrate that pretrained vision models can be leveraged to learn distortion-robust representations, which can then be effectively applied to downstream tasks operating on distorted observations.
    In particular, we propose an asymmetric knowledge distillation framework in which both teacher and student are initialized from the same pretrained Vision Transformer but receive different views of each image: the teacher processes clean images, while the student sees their distorted versions.
    We introduce multi-level distillation that aligns global embeddings, patch-level features, and attention maps and show that the student is able to approximate clean-image representations despite never directly accessing clean data.
    We evaluate our approach on image classification tasks across several datasets and under various distortions, consistently outperforming existing alternatives for the same amount of human supervision.

\end{abstract}

\section{Introduction}
Learning robust visual representations from images under noise, occlusion, blur, or other distortions remains a central challenge in computer vision.
In many real-world applications, such as medical imaging, remote sensing, and autonomous systems, high-quality, clean data is often scarce, expensive to collect, or altogether unavailable.
As a result, models must often operate directly on corrupted observations, where extracting meaningful semantic features becomes substantially more difficult.
Supervised learning methods, typically relying on clean and labeled datasets, tend to significantly degrade when trained or evaluated on such distorted inputs.

Recent advances in self-supervised vision models, particularly Vision Transformers (ViT)~\cite{dosovitskiy2021an}, have led to powerful encoders trained on massive image datasets that have demonstrated impressive performance across a wide range of image understanding tasks.
However, when applied directly to corrupted inputs, these models struggle to preserve their semantic representations, as distortions disrupt both low-level visual cues and higher-level structural information.
This raises the question: can we recover the clean representations of a pretrained encoder while only observing distorted versions of the images?
In this work, we investigate this question and show that it is possible to approximate the clean feature space of a strong pretrained encoder, producing robust representations that are useful across diverse downstream tasks which lack access to clean data.


Knowledge distillation has emerged as a standard paradigm for transferring knowledge from a teacher to a student model, often leading to improvements in model compactness, generalization, and training efficiency across various vision tasks including image classification, detection, and segmentation.
Although recent studies have begun to explore its benefits in improving model robustness, its potential for learning effective representations under severe visual distortions remains underexplored.
A notable exception is the work of Ravula et al.~\cite{ravula2021inverse}, which leverages a pretrained contrastive encoder to recover representations of corrupted images, demonstrating that pretrained models can serve as effective priors for representation recovery under severe distortions.

In this paper, we provide a simple yet effective knowledge distillation approach for training robust image encoders capable of handling highly distorted images.
In particular, we propose a ViT-based knowledge distillation framework where a pretrained teacher processes clean images while the student model observes only their distorted counterparts, addressing an asymmetric learning setting where both networks are initialized from the same pretrained Vision Transformer but observe fundamentally different views of each image.
Extending beyond contrastive representation recovery~\cite{ravula2021inverse}, we introduce a multi-level distillation objective that aligns global image embeddings, patch-level features, and attention maps, providing richer and more structured supervision that guides the student to learn distortion-invariant representations without ever observing clean data.

We assess the robustness and transferability of the learned image encoders through classification and transfer learning experiments across various datasets and distortion types, including random missing pixels, as well as Gaussian noise and blur.
We show that given the same amount of labeled data, our method consistently outperforms supervised baselines and state-of-the-art self-supervised approaches, demonstrating that robust and transferable representations can be effectively obtained from corrupted inputs.

Altogether, our main contributions are as follows:
\begin{itemize}
    \item We propose a knowledge distillation framework that leverages pretrained ViT models to train robust encoders for highly distorted images.
    \item We introduce and combine multi-level distillation objectives, namely, global embeddings, patch-level representations, and attention maps, that provide stronger supervision and improve alignment between teacher and student.
    \item We demonstrate through extensive experiments the robustness and generalization of learned encoders, achieving significantly improved results over existing baselines.
\end{itemize}

\section{Related Work}
\subsection{Robust Image Recognition}
Despite impressive performance on clean images, deep neural networks have been shown to be fragile to simple corruptions.
Early work revealed the severe degradation of image classifiers when evaluated on common perturbations such as noise, blur, and compression artifacts~\cite{dodge2016understanding,hendrycks2019benchmarking}, even after fine-tuning on these specific corruptions~\cite{dodge2017study}.
\cite{geirhos2018generalisation} further demonstrated that corruption-specific training fails to generalize to unseen corruptions.
To enable systematic evaluation, \cite{hendrycks2019benchmarking} introduced ImageNet-C, which has since become a standard benchmark for robustness to distributional shifts, spanning diverse corruption types and severity levels.

A prominent line of work addresses distortion robustness through data augmentation, ranging from simple techniques like random cropping and flipping to more sophisticated strategies.
Methods such as Mixup~\cite{zhang2018mixup} and AugMix~\cite{hendrycks2020augmix} improve generalization by interpolating between training examples or composing diverse augmentation chains, exposing models to a wider range of corruptions at training time.
Complementary approaches include feature normalization schemes and architectural design choices that promote invariance to input variations~\cite{schneider2020improving,mao2022towards}.
More recently, it has been observed that finetuning models pretrained on large-scale datasets, rather than training end-to-end from scratch, can substantially improve robustness to distribution shift~\cite{hendrycks2019usingpre,xie2020self}, indicating that rich pretrained representations carry inherent robustness benefits.
Nevertheless, simply finetuning on corrupted images remains insufficient in highly degraded regimes, motivating more principled approaches for representation learning under distortion.

\subsection{Denoising and Image Restoration Methods}
A parallel body of work approaches corrupted inputs through explicit image reconstruction, aiming to recover a clean image from its degraded observation.
Classical denoising methods~\cite{buades2011non} have been succeeded by deep learning approaches such as FFDNet~\cite{zhang2018ffdnet}, with more recent methods leveraging transformers and diffusion models for high-quality restoration~\cite{zamir2022restormer,kawar2022denoising}.
Self-supervised denoising methods such as Noise2Noise~\cite{lehtinen2018noise2noise} and Blind-Spot Networks~\cite{krull2020probabilistic} have further relaxed the requirement for clean training targets.
However, pixel-level reconstruction is an indirect objective for downstream recognition.
In contrast, we bypass pixel reconstruction entirely and operate directly in the latent space of a pretrained encoder, recovering clean semantic representations from corrupted inputs with the explicit goal of learning distortion-robust encoders for downstream tasks.

\subsection{Self-Supervised Representation Learning}
Self-supervised learning (SSL) has produced a new generation of visual encoders that learn rich semantic representations without human annotation.
Contrastive methods such as SimCLR~\cite{chen2020simple} learn representations by attracting views of the same image while repelling views of different images.
Non-contrastive approaches including BYOL~\cite{grill2020bootstrap} and SimSiam~\cite{chen2021exploring} achieve similar results without negative pairs.
Masked image modeling methods such as MAE~\cite{he2022masked} train encoders to reconstruct masked image patches, encouraging the model to capture structural and contextual information.
DINO~\cite{caron2021emerging} and DINOv2~\cite{oquab2024dinov2} apply self-distillation with multi-crop augmentation and have been shown to produce features with exceptional semantic structure.
Unlike supervised representations, features learned through self-supervised pretraining have been shown to be more robust to distribution shift~\cite{hendrycks2019usingself}, as they tend to capture high-level semantic content rather than low-level texture statistics.
Our work builds on this insight, leveraging the semantic structure captured by strong SSL pretrained Vision Transformers and adapting it to the corrupted-input setting through knowledge distillation, enabling robust downstream performance without access to clean training data.

\subsection{Knowledge Distillation Methods}
Knowledge distillation~\cite{hinton2015distilling} was originally proposed to transfer the soft output distribution of a large teacher to a compact student, providing richer training signal than hard labels alone.
Subsequent work has extended distillation beyond the output layer to intermediate feature representations.
FitNets~\cite{romero2015fitnets} align intermediate hidden layers, while methods such as RKD~\cite{park2019relational} and CRD~\cite{tian2020contrastive} transfer relational and contrastive structures from teacher to student.
Attention transfer~\cite{zagoruyko2017paying} explicitly aligns the attention maps of teacher and student networks, encouraging similar spatial focus.
More recently, distillation has been adopted in the context of self-supervised learning, where DINO~\cite{caron2021emerging} can be interpreted as a form of self-distillation, and methods like DeiT~\cite{touvron2021training} integrate distillation tokens to leverage powerful teacher models during ViT training.

The most closely related work to ours is that of Ravula et al.~\cite{ravula2021inverse}, which leverages a CLIP-pretrained~\cite{radford2021learning} encoder to recover representations of highly corrupted images through a contrastive objective.
Our work builds on these directions and extends them by adopting a ViT-based teacher--student framework with a multi-level distillation objective that aligns global embeddings, patch-level features, and attention maps, providing richer supervision and
guiding the student to bridge the distribution gap between corrupted observations and clean semantic representations.


\section{Method}
Our goal is to learn visual representations that remain semantically meaningful under severe image corruptions.
To this end, we train a robust encoder via a teacher--student distillation framework in which a pretrained model operating on clean images supervises a student model receiving distorted inputs.
We frame robustness as a \emph{representation recovery} problem in latent space: rather than reconstructing clean pixels, the student learns to infer feature representations semantically equivalent to those of its clean counterpart.

\subsection{Problem Formulation}
Let $x \in \mathbb{R}^{H \times W \times 3}$ denote a clean image and $\tilde{x} = \mathcal{D}(x)$ its corrupted counterpart, where $\mathcal{D}$ models a known distortion process (e.g., additive noise, blur, or masking).
Given a pretrained encoder $f_\phi$ serving as a semantic reference, we seek a robust encoder $f_\theta^{\mathcal{D}}$ satisfying:
\begin{equation}
    f_\theta^{\mathcal{D}}(\tilde{x}) \approx f_\phi(x),
\end{equation}
where each $f_\theta^{\mathcal{D}}$ is trained exclusively on inputs corrupted by $\mathcal{D}$, allowing it to specialize toward that corruption regime rather than averaging across distortions.
Operating in the lower-dimensional, semantically structured feature space of $f_\phi$ promotes invariance to low-level corruptions while preserving the high-level content required for downstream tasks such as classification and transfer learning.

\subsection{Knowledge Distillation Framework}
We adopt a teacher--student distillation setup where both models are initialized from the same pretrained ViT encoder checkpoint.
The teacher $f_\phi$ processes clean images; the student $f_\theta$ processes their distorted counterparts.
The teacher remains frozen throughout training, providing a stable semantic anchor, while only the student parameters $\theta$ are updated via backpropagation.
We note that no class labels are used during distillation, as supervision derives entirely from the teacher's internal representations, making our approach fully self-supervised.

Given a clean--distorted pair $(x, \tilde{x})$, we extract from each model the full sequence of token embeddings after the last transformer layer:
\begin{equation}
    H_T = f_\phi(x) = \bigl[h_T^{\mathrm{cls}},\, h_T^{1},\, \dots,\, h_T^{P}\bigr] \in \mathbb{R}^{(P+1) \times d},
\end{equation}
\begin{equation}
    H_S = f_\theta(\tilde{x}) = \bigl[h_S^{\mathrm{cls}},\, h_S^{1},\, \dots,\, h_S^{P}\bigr] \in \mathbb{R}^{(P+1) \times d},
\end{equation}
where $h^{\mathrm{cls}}$ is the class token encoding global image semantics, $\{h^{p}\}_{p=1}^{P}$ are patch tokens encoding spatially localized features, and $d$ denotes the embedding dimensionality of the ViT encoder.

\begin{figure}[h]
    \centering
    \includegraphics[width=\textwidth]{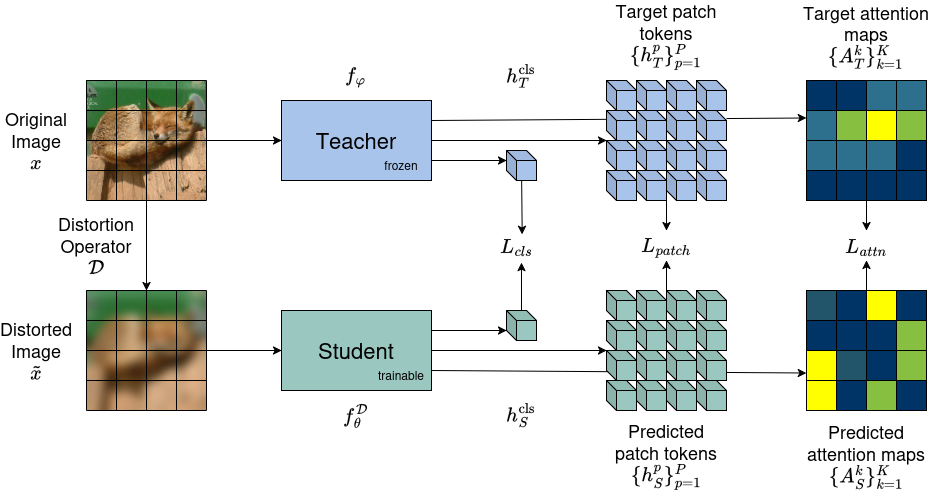}
    \caption{Overview of our multi-target distillation framework.
    A frozen teacher ViT processes clean images while a student ViT of identical architecture receives their distorted counterparts, being trained to align with the teacher at three complementary levels: global embeddings, patch-level features, and attention maps. In this way, the student learns to recover clean semantic representations directly from corrupted inputs, producing robust encoders that are effective for downstream tasks on distorted images.
    }
    \label{fig:method}
\end{figure}

\subsection{Multi-Target Distillation Objective}
To effectively transfer knowledge from teacher to student, we introduce a \emph{multi-target distillation objective} that aligns three complementary levels of the ViT representation: global semantics, local spatial structure, and relational attention patterns.
Jointly distilling these hierarchically distinct signals encourages the student to recover a representation that faithfully reflects the semantic content of the clean image despite input corruptions.

\subsubsection{Global Semantic Alignment}
The class token $h^{\mathrm{cls}}$ aggregates context across all patches into an image-level representation.
We align the teacher and student class tokens via MSE:
\begin{equation}
    \mathcal{L}_{\mathrm{cls}} = \bigl\| h_S^{\mathrm{cls}} - h_T^{\mathrm{cls}} \bigr\|_2^2.
\end{equation}
This loss directly encourages the student to produce globally coherent representations that preserve the categorical and semantic content of the clean image.

\subsubsection{Local Spatial Alignment}
Patch tokens encode fine-grained visual details and spatial structure.
To maintain local semantic consistency under distortions, we align patch embeddings at each position:
\begin{equation}
    \mathcal{L}_{\mathrm{patch}} = \frac{1}{P} \sum_{p=1}^{P} \bigl\| h_S^{p} - h_T^{p} \bigr\|_2^2.
\end{equation}
This objective prevents corruptions from collapsing local spatial structure into globally averaged features, preserving the full image layout.

\subsubsection{Attention Structure Alignment}
Beyond feature activations, we distill the \emph{relational structure} encoded by self-attention, specifically how the model distributes focus over spatial regions when forming the global class token representation.
For each attention head $k \in \{1,\dots,K\}$ in the last transformer layer, we extract the attention weights $A_T^k, A_S^k \in \Delta^{P}$ of the class token attending to all patch tokens.
Interpreting these as probability distributions via softmax with temperature $\tau$, we minimize their KL divergence:
\begin{equation}
    \mathcal{L}_{\mathrm{attn}} = \frac{1}{K} \sum_{k=1}^{K} \mathrm{KL}\!\left(\, \sigma(A_T^k / \tau) \,\Big\|\, \sigma(A_S^k / \tau) \right),
\end{equation}
where $\sigma(\cdot)$ denotes the softmax function and $\tau{=}2$ is a temperature parameter that smooths the distributions, preventing the loss from being dominated by the most attended patches.
We distill each head independently rather than averaging across heads, preserving the heterogeneous relational cues captured by different attention mechanisms.

\subsubsection{Overall Objective}
The three components are combined into a unified distillation loss:
\begin{equation}
    \mathcal{L} = \lambda_{\mathrm{cls}}\,\mathcal{L}_{\mathrm{cls}} +
                  \lambda_{\mathrm{patch}}\,\mathcal{L}_{\mathrm{patch}} +
                  \lambda_{\mathrm{attn}}\,\mathcal{L}_{\mathrm{attn}}.
\end{equation}
Since $\mathcal{L}_{\mathrm{attn}}$ operates on probability distributions and is therefore of smaller magnitude than the MSE-based terms, $\lambda_{\mathrm{attn}}$ is set to a larger value to ensure all three loss components contribute comparably to the total gradient.
We set $\lambda_{\mathrm{cls}}{=}1$, $\lambda_{\mathrm{patch}}{=}1$, and $\lambda_{\mathrm{attn}}{=}50$, which we found to work best in practice.

Figure~\ref{fig:method} provides an overview of the proposed approach, illustrating the teacher--student setup and the three levels of representation alignment.

\subsection{Training Details}
\label{sec:training_details}
We train our robust encoders on ImageNet-100, a 100-class subset of ImageNet~\cite{russakovsky2015imagenet} containing approximately 130K training images, providing sufficient semantic diversity while substantially reducing computation relative to the full benchmark.
Following~\cite{tian2020contrastive}, we extract the exact same classes from the original set, ensuring reproducibility and consistency with prior work.
We initialize teacher and student models from the same pretrained DINO-ViT-B/16~\cite{caron2021emerging} checkpoint, chosen for its strong visual representations, well-structured feature space, and interpretable attention maps.

Input images are resized to $224{\times}224$ and augmented with random resized cropping and horizontal flipping.
We generate their corrupted counterparts by applying image distortions on-the-fly, spanning random pixel masking, additive Gaussian noise, and Gaussian blur.

Models are trained for 25 epochs with a batch size of 128 using AdamW with a peak learning rate of $10^{-5}$, cosine annealing, and weight decay $10^{-4}$, in mixed-precision (float16).
All experiments are conducted on 2 NVIDIA RTX A6000 GPUs.

\section{Experiments}
This section empirically demonstrates the effectiveness and robustness of our asymmetric knowledge distillation framework.
We first evaluate performance on the same dataset used to pretrain the backbone, enabling a controlled assessment of distortion robustness without domain shift.
We then study cross-dataset transfer to measure the generalization capability of the learned representations.
Finally, we conduct extensive ablation studies and analytical experiments to disentangle the contributions of each distillation objective and to better understand the source of the robustness improvements.

\subsection{Datasets}
We evaluate our method on one in-domain benchmark and several transfer datasets spanning natural, remote sensing, and medical imagery.

For in-domain evaluation, we use ImageNet-100, the same dataset used for pretraining our encoders as described in Sec.~\ref{sec:training_details}, with 5K validation images held out for evaluation.
For transfer learning, we employ CIFAR-10 and CIFAR-100, each consisting of 60,000 32×32 images (50K train, 10K test), with 10 and 100 classes respectively.
We further evaluate on STL-10, which contains 5,000 training and 8,000 test images at a higher 96×96 resolution and across 10 classes.
To assess cross-domain generalization, we use RESISC45, comprising 31,500 aerial images from 45 scene categories (700 images per class), and CAMELYON17, a histopathology dataset with over 450K image patches collected from five medical centers for tumor classification.
These datasets enable evaluation of robustness both within the pretraining domain and under substantial distribution shifts.

\subsection{Distortions}
\label{sec:distortions}
Following~\cite{ravula2021inverse}, we consider three families of synthetic image corruption, each instantiated at two severity levels.

\textbf{Pixel masking} randomly zeros out a fixed fraction of pixels, with masking ratios of 0.75 and 0.90 (retaining 25\% and 10\% of pixel values, respectively).
\textbf{Gaussian noise} adds zero-mean i.i.d. noise with standard deviation $\sigma \in \{0.3, 0.5\}$, clipped to the valid intensity range.
\textbf{Gaussian blur} applies an isotropic kernel with (kernel size, $\sigma$) pairs of (21, 5) and (37, 9), respectively.

These three corruption families span distinct degradation axes: severe information removal, stochastic intensity perturbation, and suppression of high-frequency content.

\subsection{Methods}
In our evaluation, we compare the proposed asymmetric multi-level distillation framework against two baselines for learning under severe image distortions. We benchmark our method and alternative approaches on image classification tasks across several datasets and distortion types.

\subsubsection{Supervised Baseline}
As a direct baseline, we fine-tune a pretrained Vision Transformer on the distorted training images using standard supervised learning.
The model is initialized from the same pretrained backbone as our teacher and student networks, and all parameters are updated during fine-tuning.
This baseline assesses whether robustness can be achieved solely through end-to-end supervised adaptation to corrupted inputs, without any explicit alignment to clean representations.
It also controls for the effect of initialization and model capacity, isolating the benefit of our distillation strategy.

\subsubsection{Contrastive Inversion (CI)}
We further compare against the method proposed in \cite{ravula2021inverse}, which exploits pretrained contrastive encoders adapting them for extracting robust representations from highly distorted inputs.
The approach builds upon a CLIP-pretrained ResNet-101~\cite{he2016deep} and leverages its contrastive representation space as a robust prior.
In this framework, corrupted observations are mapped into the pretrained feature space, guided by a contrastive loss that pulls student representations toward the corresponding teacher representations and pushes apart representations of different images.
In our classification setting, the resulting student encoder is subsequently fine-tuned on distorted training images using supervision.

\subsection{In-Domain Experiments}
In this section, we assess the robustness of the learned image encoders on classification under severe distortions.
After our distillation pretraining, we fine-tune the student models in a supervised manner using the labels of the ImageNet-100 training set.
We then report top-1 classification accuracy on the corresponding validation set, applying the same distortion type and severity during both training and evaluation.

This setting evaluates the ability of our models to extract semantic representations when only degraded inputs are available.
Since both pretraining and supervised fine-tuning are conducted on the same data distribution, this constitutes an in-domain evaluation.
All methods are compared across the distortion types and severity levels described in Sec.~\ref{sec:distortions}.

The results are presented in Table~\ref{tab:imagenet100}.
Our method achieves the highest top-1 accuracy in most distortion settings, particularly under Gaussian noise and blur, where it consistently outperforms both baselines.
Under random pixel masking, our approach remains competitive, achieving performance close to the best one (CI), while maintaining clear improvements over the supervised baseline.

\begin{table}[h]
    \caption{Top-1 classification accuracy (\%) on ImageNet-100 under different distortion types and severity levels (in-domain evaluation).
    Results are reported as mean $\pm$ standard deviation over 10 random initializations of distortions in the validation set.
    }
    \label{tab:imagenet100}
    \centering
    \begin{tabular}{@{}lccc@{}}
        \toprule
        Distortion & CI~\cite{ravula2021inverse} & Supervised Baseline & Ours\\
        \midrule
        Random Mask $75\%$ & $\mathbf{86.27\pm0.18}$ & $82.90\pm0.25$ & $85.41\pm0.28$ \\
        Random Mask $90\%$ & $\mathbf{83.92\pm0.23}$ & $78.26\pm0.26$ & $81.94\pm0.26$\\
        Gaussian Noise $\sigma=0.3$ & $83.04\pm0.21$ & $87.48\pm0.22$ & $\mathbf{88.77\pm0.10}$\\
        Gaussian Noise $\sigma=0.5$ & $77.49\pm0.37$ & $82.56\pm0.31$ & $\mathbf{84.64\pm0.24}$\\
        Gaussian Blur $n=21$ & $84.72$ & $87.46$ & $\mathbf{89.11}$\\
        Gaussian Blur $n=37$ & $78.68$ & $80.08$ & $\mathbf{83.94}$\\
    \bottomrule
    \end{tabular}
\end{table}

\subsection{Transfer Learning}
We next evaluate whether robustness learned through distortion-aware pretraining on ImageNet-100 transfers to other datasets.
The pretrained student encoders are fine-tuned on downstream classification benchmarks using their respective training splits.
For each corruption family, we consider only its most severe configuration and transfer the corresponding pretrained model.
During fine-tuning and evaluation, images are corrupted using the same distortion type as in pretraining.
Performance is reported as top-1 classification accuracy on  
test sets.

The transfer benchmarks include CIFAR-10, CIFAR-100, STL-10, RESISC45, and CAMELYON17, covering natural images, aerial scenes, and medical imagery.
These datasets introduce varying degrees of distribution shift relative to ImageNet-100, allowing us to assess the generalization of the learned image encoders under degraded inputs.
All methods are trained and evaluated under the same protocol for fair comparison.

The transfer learning results are shown in Table~\ref{tab:transfer}.
Our method consistently achieves the highest accuracy across all datasets and distortion types.
In most cases, it provides substantial improvements over both baseline approaches, particularly on RESISC45 and CAMELYON17, which exhibit a pronounced domain shift relative to ImageNet-100.
These findings demonstrate that the proposed distortion-aware pretraining yields robust and transferable encoders across diverse tasks and input degradations.

\begin{table}[h]
    \caption{Transfer learning under severe image distortions.
    Top-1 classification accuracy (\%) on five downstream datasets under three corruption types (RM: 90\% random masking; GN: Gaussian noise with $\sigma=0.5$; GB: Gaussian blur with kernel size $37$ and $\sigma=9$). Results are reported as mean $\pm$ standard deviation over 10 random realizations of the distortions applied to the test images.
    }
    \label{tab:transfer}
    \centering
    \begin{tabular}{@{}lccccc@{}}
        \toprule
        Model & CIFAR-10 & CIFAR-100 & STL-10 & RESISC45 & CAMELYON17 \\
        \midrule
        CI~\cite{ravula2021inverse} (RM) & $93.14\pm0.14$ & $65.21\pm0.15$ & $86.90\pm0.21$ & $79.79\pm0.26$ & $55.84\pm0.04$ \\
        Baseline (RM) & $90.69\pm0.15$ & $65.76\pm0.24$ & $69.51\pm0.17$ & $80.88\pm0.38$ & $56.20\pm0.06$ \\
        Ours (RM) & $\mathbf{95.05\pm0.10}$ & $\mathbf{75.18\pm0.12}$ & $\mathbf{90.99\pm0.09}$ & $\mathbf{83.75\pm0.22}$ & $\mathbf{76.03\pm0.05}$ \\
        \midrule
        CI~\cite{ravula2021inverse} (GN) & $86.46\pm0.24$ & $54.03\pm0.15$ & $78.58\pm0.26$ & $68.11\pm0.33$ & $62.27\pm0.05$ \\
        Baseline (GN) & $91.86\pm0.12$ & $70.02\pm0.28$ & $79.95\pm0.23$ & $81.21\pm0.22$ & $55.86\pm0.04$ \\
        Ours (GN) & $\mathbf{94.72\pm0.19}$ & $\mathbf{74.58\pm0.28}$ & $\mathbf{92.11\pm0.25}$ & $\mathbf{83.53\pm0.26}$ & $\mathbf{65.24\pm0.07}$ \\
        \midrule
        CI~\cite{ravula2021inverse} (GB) & $92.21$ & $63.55$ & $80.17$ & $74.60$ & $73.88$ \\
        Baseline (GB) & $95.91$ & $80.56$ & $86.22$ & $84.68$ & $59.73$ \\
        Ours (GB) & $\mathbf{97.02}$ & $\mathbf{82.19}$ & $\mathbf{90.35}$ & $\mathbf{86.03}$ & $\mathbf{74.77}$ \\
    \bottomrule
    \end{tabular}
\end{table}

\subsection{Ablation on Distillation Loss Components}
To better understand the contribution of the different components of our distillation objective, we perform an ablation study where we selectively remove alignment terms during pretraining.
Specifically, we evaluate variants that use only the global feature alignment, combinations of global and local alignment, and global with attention alignment.
After pretraining, all variants are fine-tuned and evaluated on ImageNet-100 under the same distortion settings as in the main experiments.

The results are summarized in Table~\ref{tab:ablation}.
While global alignment alone already provides strong performance, incorporating additional alignment constraints consistently improves the robustness of the learned representations.
In particular, adding local alignment yields noticeable gains under both random masking and Gaussian noise.
The full objective, combining global, local, and attention alignment, achieves the best performance across all distortion types, indicating that these components provide complementary supervision during distillation.

\begin{table}[t]
    \caption{Ablation of distillation loss components.
    Top-1 classification accuracy (\%) on ImageNet-100 under severe distortions when removing alignment terms from the distillation objective. Global: global embedding; Local: patch-level; Attention: attention map alignment. Results are mean $\pm$ standard deviation over 10 random distortion realizations on the validation set.
    }
    \label{tab:ablation}
    \centering
    \begin{tabular}{@{}lccc@{}}
        \toprule
        & Random Mask & Gaussian Noise & Gaussian Blur \\
        Model & $90\%$ & $\sigma=0.5$ & $n=37$\\
        \midrule
        Global & $81.21\pm0.41$ & $84.05\pm0.23$ & $83.56$ \\
        Gl. $+$ Local & $81.85\pm0.35$ & $84.44\pm0.28$ & $83.74$ \\
        Gl. $+$ Attention & $81.39\pm0.29$ & $83.93\pm0.36$ & $83.82$ \\
        Gl. $+$ Local $+$ Attention & $81.94\pm0.26$ & $84.64\pm0.24$ & $83.94$ \\
        \bottomrule
    \end{tabular}
\end{table}

\subsection{Robustness Under Increasing Distortions}
We further evaluate robustness by testing the models on ImageNet-100 validation images corrupted with distortions that are more severe than those encountered during training.
Specifically, we increase the distortion strength for random masking, Gaussian noise, and Gaussian blur beyond the levels used during pretraining and fine-tuning.
This experiment assesses how well the learned representations generalize to degradations outside the training distribution.

\begin{figure}[h]
    \centering
    \includegraphics[width=\textwidth]{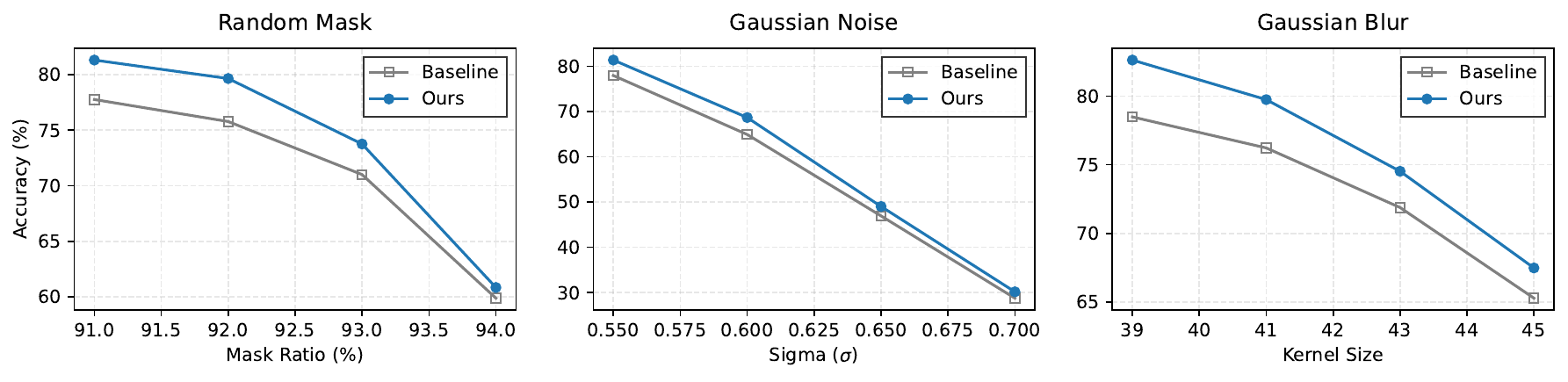}
    \caption{Robustness under increasing distortions.
    Top-1 classification accuracy (\%) on the ImageNet-100 validation set as the distortion intensity exceeds the levels used during training. Models are fine-tuned on random masking (90\%), Gaussian noise ($\sigma=0.5$), and Gaussian blur (kernel size $37$), respectively.
    }
    \label{fig:robustness}
\end{figure}

The results are shown in Fig.~\ref{fig:robustness}.
Across all distortion types, our method consistently outperforms the supervised baseline as the corruption severity increases.
While accuracy degrades for both models under stronger distortions, our approach maintains a clear advantage, indicating that our distillation strategy leads to representations that remain more stable and informative under severe input perturbations.

\subsection{Label-Efficient Fine-Tuning}
We also investigate the performance of our distilled representations when only a fraction of training labels is available.
For each distortion type i.e., random masking (90\%), Gaussian noise ($\sigma=0.5$), and Gaussian blur (kernel size $37$), both our student model and the supervised baseline are fine-tuned on subsets of the ImageNet-100 training set ranging from 0.5\% to 100\% of the labels.
Models are evaluated on the full validation set under the same distortions.

\begin{figure}[h]
    \centering
    \includegraphics[width=\textwidth]{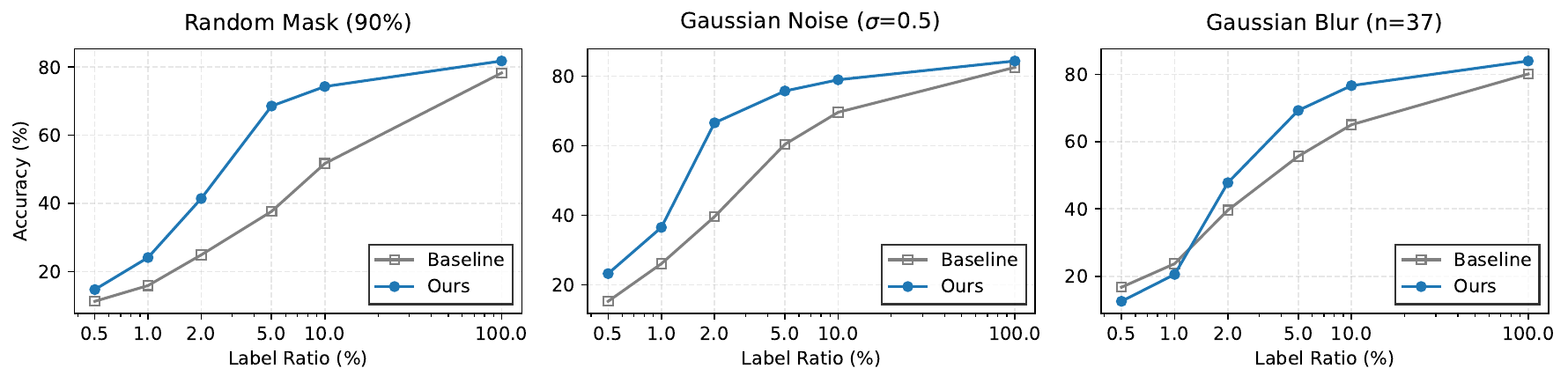}
    \caption{Label-efficient performance under distortions.
    Top-1 classification accuracy (\%) on ImageNet-100 validation images for varying fractions of training labels (log-scale x-axis).
    }
    \label{fig:label_efficiency}
\end{figure}

As shown in Fig.~\ref{fig:label_efficiency}, our method consistently outperforms the supervised baseline across almost all distortions and label fractions.
The advantage is most pronounced at very low label ratios, indicating that the distillation pretraining produces representations that are both label-efficient and robust to severe input corruptions.
Even with only 5–10\% of labels, performance approaches that of the fully supervised model trained on all labels.

\subsection{Attention Map Analysis}
\begin{figure}[t]
    \centering
    \includegraphics[width=\textwidth]{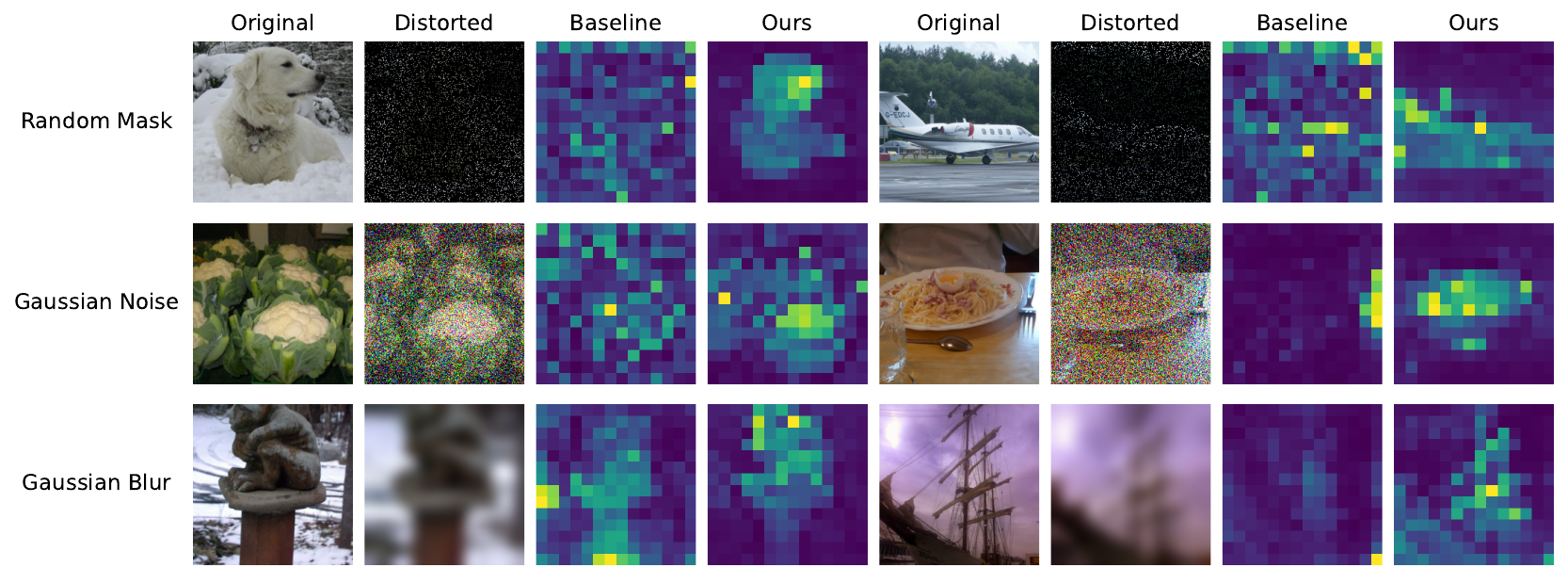}
    \caption{Attention maps of our distilled student models and the supervised baselines on ImageNet-100 validation images, under different distortion types.
    For each example, we show the original image, the distorted input, and the attention maps predicted by the supervised baseline and the encoder trained by our method.
    }
    \label{fig:attention_maps}
\end{figure}

Figure~\ref{fig:attention_maps} visualizes the attention maps produced by our distilled student models and the supervised baselines on distorted images.
Despite never being exposed to class labels during pretraining, our models consistently focus on semantically relevant regions, correctly identifying the objects of interest even under severe distortions.
In contrast, the attention maps of the supervised baselines are largely deteriorated, failing to localize meaningful structures in most cases.
Beyond confirming that our distillation approach produces encoders that remain semantically coherent under severe input corruptions, these results also highlight the interpretability of the learned representations, as the attended regions remain meaningful and spatially consistent with the clean image semantics.

\section{Discussion}
In this work, we address distortion-robust visual representation learning by framing it as a clean representation recovery task.
Rather than reconstructing clean pixels or relying on large-scale contrastive pretraining, we supervise the student directly in the teacher's semantic feature space.
We show this to be both effective and practical, as even under 90\% pixel masking or strong Gaussian noise and blur, the student recovers representations that closely approximate those of the clean-image teacher, yielding strong downstream performance.

Our multi-level distillation objective aligns student and teacher at three levels of representation.
The ablation study confirms that each component contributes distinct information, where global alignment provides a strong initial point, and incorporating local patch alignment adds consistent gains across all distortion types.
Attention alignment provides further improvements, particularly under masking and blur, with the full objective outperforming all partial combinations, thus validating that these signals are complementary.

A notable finding is that our method, leveraging a DINO-ViT-B/16 pretrained on ImageNet-1K, consistently matches or surpasses CI across most distortion types and substantially outperforms it on transfer benchmarks, despite CI relying on a CLIP backbone trained on hundreds of millions of image-text pairs.
This suggests that DINO's self-supervised objective produces semantically rich, spatially structured features that are particularly well-suited as distillation targets, whereas CLIP's contrastive objective is optimized for cross-modal alignment rather than fine-grained visual structure.
The supervised baseline, sharing the same DINO backbone, also underperforms our method, confirming that our gains come from the explicit alignment to clean representations, and not from the pretrained initialization alone.

The absence of label supervision during the distillation pretraining is practically significant.
In domains where distortion robustness matters most such as medical imaging and remote sensing, annotations are often scarce and costly.
Our label-efficiency experiments show that with as few as 5\% of training labels, our pretrained encoders approach the performance of a fully supervised model, confirming that the distilled representations are semantically well-organized prior to any labeled fine-tuning.

Finally, a current limitation of our work is that a separate encoder is trained per distortion type.
Handling unknown or mixed corruptions within a single model remains a direction for future work.

\section{Conclusion}
In this paper, we frame distortion-robust visual representation learning as a clean representation recovery problem, where a model must infer semantically meaningful features from highly distorted observations.
We present a self-supervised distillation approach leveraging pretrained Vision Transformers that aligns a student encoder, which never observes clean images, with a frozen teacher at multiple levels of representation, namely global embeddings, patch features, and attention maps.
Experiments across diverse datasets and distortion types demonstrate consistent improvements over supervised fine-tuning and a contrastively pretrained baseline, with significant gains in label-scarce and domain-shifted settings, suggesting multi-level distillation as an effective strategy for robust, transferable visual representation learning.


\begin{credits}

\subsubsection{\discintname}
The authors have no competing interests to declare that are
relevant to the content of this article. 
\end{credits}
%
%
%
\bibliographystyle{splncs04}
\bibliography{references}
%




\end{document}